\documentclass[letterpaper, 10 pt, conference]{ieeeconf}  

\IEEEoverridecommandlockouts                              

\usepackage{graphicx} 
\usepackage{amsmath} 
\usepackage{multirow}
\usepackage{multicol}
\usepackage{diagbox}
\title{\LARGE \bf
HITL-D: Human In The Loop Diffusion Assisted Shared Control
}

\author{Riley Zilka$^{*}$ and Sergey Khlynovskiy$^{*}$ and Allie Wang$^{*}$ and Martin Jagersand$^{*}$
\thanks{$^{*}$Department of Computing Science,
        University of Alberta, Edmonton Alberta, Canada, T6G 2E8.
        {\tt\small \{rzilka, khlynovs, luo3, mj7\}@ualberta.ca}}%
}

\begin{document}

\maketitle
\thispagestyle{empty}
\pagestyle{empty}

\begin{abstract}

Autonomous manipulation systems have achieved remarkable capabilities, yet the integration of human expertise with diffusion-based policies in shared control remains relatively unexplored. In this paper, we propose Human-In-The-Loop Diffusion (HITL-D), a shared control framework that enhances user performance in multi-step, insertion, and fine manipulation tasks. HITL-D leverages a novel combination of diffusion-based policies and human control to provide autonomous end effector orientation updates conditioned on a scene point cloud and the Cartesian position of the end effector. This approach reduces the number of joystick control axes required, thereby lowering mental workload. In a multi-task user study with 12 participants, HITL-D reduced average task completion times by 40\%, decreased perceived workload by 37\%, and improved Likert-scale ratings for independence, intuitiveness, and confidence compared to traditional teleoperation methods. These results demonstrate that HITL-D effectively integrates human expertise with autonomous assistance, improving both objective and subjective aspects of teleoperation.

\end{abstract}

\section{INTRODUCTION}

Wheelchair-mounted robotic arms (WMRAs) hold significant promise in enhancing independence and quality of life for individuals with motor impairments \cite{driessen2001manus, maheu2011evaluation}. These systems extend the functional capabilities of powered wheelchairs by enabling users to perform activities of daily living (ADLs) such as grasping objects, opening doors, or manipulating devices \cite{9811960}. However, despite decades of research and technological advancement, WMRAs remain underutilized in practice due to barriers such as cost and accessibility \cite{bourassa2023wheelchair}. A major barrier is the difficulty of control without training. Operating a high-degree-of-freedom robotic arm through limited interfaces, such as a standard 2 or 3 degree of freedom wheelchair joystick, imposes substantial mental workload on the user \cite{herlant2016assistive}. In the United States alone, more than 9 million people live with a self-care disability \cite{cdcdisability}. Therefore, developing new systems to reduce mental workload and effort during operation has the potential to improve the quality of life for millions of people.

Shared control is a popular field aiming to reduce workload, improve the ease of control, and minimize degrees of control input. It aims to solve tasks that could be done individually by both the robot and a human, but in such a way that the autonomous and manual inputs from both parties are blended. This reduces the effort by decreasing the necessary dimensions of control and mental effort associated with traditional methods such as Cartesian control \cite{herlant2016assistive, tijsma2005evaluation, campeau2018intuitive}. 

Shared control has evolved over several decades, starting with early supervisory frameworks in teleoperation \cite{sheridan1992telerobotics}, extending to arbitration strategies for assistive WMRAs prioritizing human input in the control and removing complete autonomy \cite{carlson2008human}, and maturing into probabilistic and intent-aware models \cite{dragan2013policy, javdani2015shared}. More recent approaches increasingly leverage learning-based methods for personalization and high-dimensional control \cite{muelling2017autonomy}. Learning-based methods are particularly justified by their ability to adapt to user preferences and environment changes while providing autonomous feedback to the system \cite{broad2018learning}. However, restricting user input, finding efficient mappings, and providing autonomous control can sometimes result in a perceived loss of authority over control \cite{javdani2015shared}. 

\begin{figure}[t]
    \centering
    \includegraphics[height=9cm,keepaspectratio]{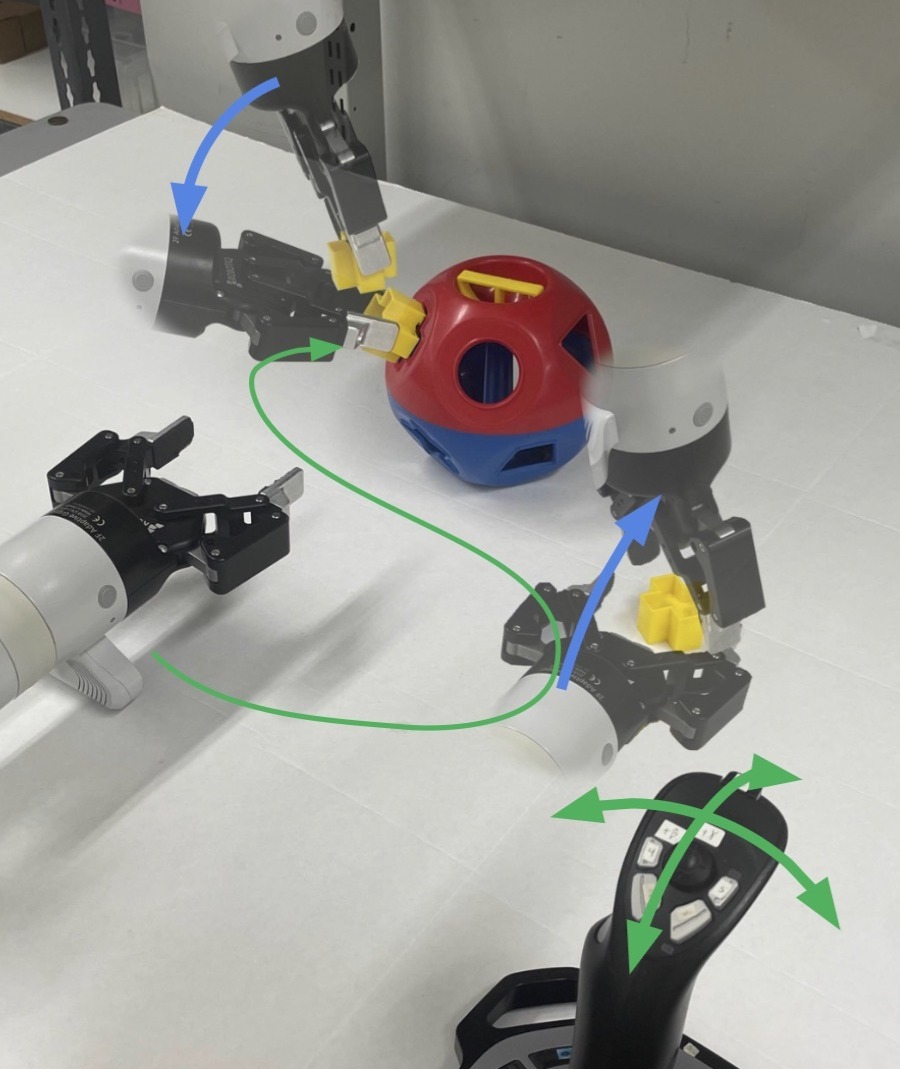}
    \caption{Our method translates the end effector (green arrow) with 3 joystick-controlled axes, while our shared-control diffusion integration continuously adjusts orientation (blue) based on demonstrations.}
    \label{title-image}
\end{figure}

Motivated by the goal of keeping the user in authority at all times while still providing useful autonomous assistance and reducing workload, we present \textbf{Human-In-The-Loop Diffusion (HITL-D)}. HITL-D is a simple and effective shared control framework that leverages expert human knowledge and task context, while combining them with the capacity of diffusion policies to represent diverse and complex robot behaviors in a flexible and generalizable format \cite{chi2023diffusion, wang2022diffusion}. Similar to the control paradigm in \cite{Jin2020cvorientationcontrol, Sun2021assymetricorientationcontrol, fu2025tasc} HITL-D works by splitting the two modes of Cartesian control, the current control standard in WMRAs: leaving the translation mode (x, y, and z) in the users control while the diffusion policy dictates orientation (yaw, pitch and roll). 

In summary, our shared control framework has the following contributions:
\begin{enumerate}
    \item We propose a novel combination of human input and the benefits of diffusion policies for our shared control framework, HITL-D.
    \item We show that HITL-D is at least as effective as traditional teleoperation methods.
    \item We show that HITL-D maintains human autonomy and preserves user perceptions of independence.
    \item We conduct a user study providing evidence that, beyond improved success rates, HITL-D reduces perceived workload compared with traditional teleoperation.
\end{enumerate}

\section{RELATED WORK}

\subsection{Shared Control}
Shared control has been explored in many applications from automotive, medical and assistive robotics \cite{Abbink2018sharedcontrol}. Providing autonomous assistance to an individual during a task while leveraging their intelligent reasoning ability creates a highly effective control paradigm. Many common applications of machine learning models in shared control include path planning \cite{bevilacqua2016path, quere2020shared}, intent detection \cite{javdani2015shared, erden2010human, jain2019probabilistic} and obstacle avoidance \cite{quere2020shared, storms2017shared}. Beyond solving task constraints, some shared control methods aim to reduce the degrees of freedom the human must control, eliminating cumbersome and sometimes frustrating mode switches \cite{przystupa2024learning}. \cite{przystupa2024learning} addresses this problem by learning a state conditioned linear mapping through multiple demonstrations to control a 7 DoF robot arm with a 2 DoF joystick controller.

This work does not address motion constraints, including intent detection or object avoidance. There is also no attempt to reduce the robots' degrees of freedom through a reduction mapping. Instead, the user controls the robot's position at all times while a diffusion policy reactively predicts the desired orientation for the task. This approach focuses on maximizing user independence while using shared autonomy to complete tasks faster and with reduced workload.

Similar approaches have addressed orientation control while leaving Cartesian position to the human operator \cite{Jin2020cvorientationcontrol, Sun2021assymetricorientationcontrol}. In \cite{Jin2020cvorientationcontrol}, control authority is swapped between the human and the autonomous system, whereas our method keeps human authority at all times, with a diffusion policy providing assistance only. \cite{Sun2021assymetricorientationcontrol} combines position and autonomous orientation using guided trajectories from keypoint poses whereas our method preserves the natural mapping of user inputs across all axes. A recent work \cite{fu2025tasc} explores the same orientation–translation split but leverages foundation models and a grasp planner for zero-shot shared control. In contrast, we train a generative policy conditioned on the operator’s translational input and a point cloud, producing orientation corrections that emerge from learned motion distributions without explicit semantic reasoning. 

\subsection{Behavior Cloning and Grasp Pose Estimation}

Behavior cloning (BC) methods \cite{torabi2018behavioral, mandlekar2021matters, codevilla2019exploring} learn policies directly from demonstrations and have been widely applied to robotic manipulation. However, standard BC typically learns a unimodal mapping from observations to actions ($o \to a$). In shared control settings where behavior is often multi-modal, a single user input could validly lead to multiple distinct manipulation strategies; common BC models often average these modes and suffer from multi-valued function discontinuities \cite{florence2022implicit}, resulting in physically inconsistent or ineffective trajectories. 

Although grasp pose estimation \cite{mahler2017dex, ten2017grasp} successfully predicts stable end-state orientations, these static goals often conflict with the user's immediate control priors. Even when providing analytically stable poses, such rigid assistance can diminish the user's sense of agency by overriding their intended manipulation strategy \cite{collier2025sense}.

\subsection{Diffusion Models for Robotics}

Diffusion models are an increasingly popular class of generative model in imitation learning and work by treating policy generation as an iterative denoising process \cite{ho2020denoising}. Unlike traditional BC, which often predicts a mean trajectory from its learned unimodal distribution, diffusion policies sample from the joint distribution of entire trajectories, $p(\tau)$. This global perspective provides two critical advantages for shared control: task multi-modality \cite{ze20243d, chi2023diffusion} and the ability to seamlessly condition on high-dimensional data \cite{wang2024conditional, zhan2024conditional}. There have been recent advancements conditioning on 3D data in diffusion \cite{ze20243d, ke20243d} demonstrating the advantages of including higher dimensional spatial information when conditioning. \cite{ze20243d} further demonstrated that these models exhibit superior generalization in high-dimensional spaces compared to standard BC.

While a simple BC model lacks the multi-modal context to fill holes in a trajectory while maintaining temporal coherence, the diffusion process iteratively refines the motion sequence to be consistent with both the learned demonstration experience and the user's real-time guidance. This ensures the system provides orientation assistance that is fluid and goal-oriented. By integrating human input directly into the control loop, we transform the model from an autonomous agent into a collaborative tool that reduces operator workload without sacrificing the expressive capacity of the underlying generative model.

\section{METHODS}

\begin{figure*}[t]
    \centering
    \includegraphics[width=1.0\textwidth]{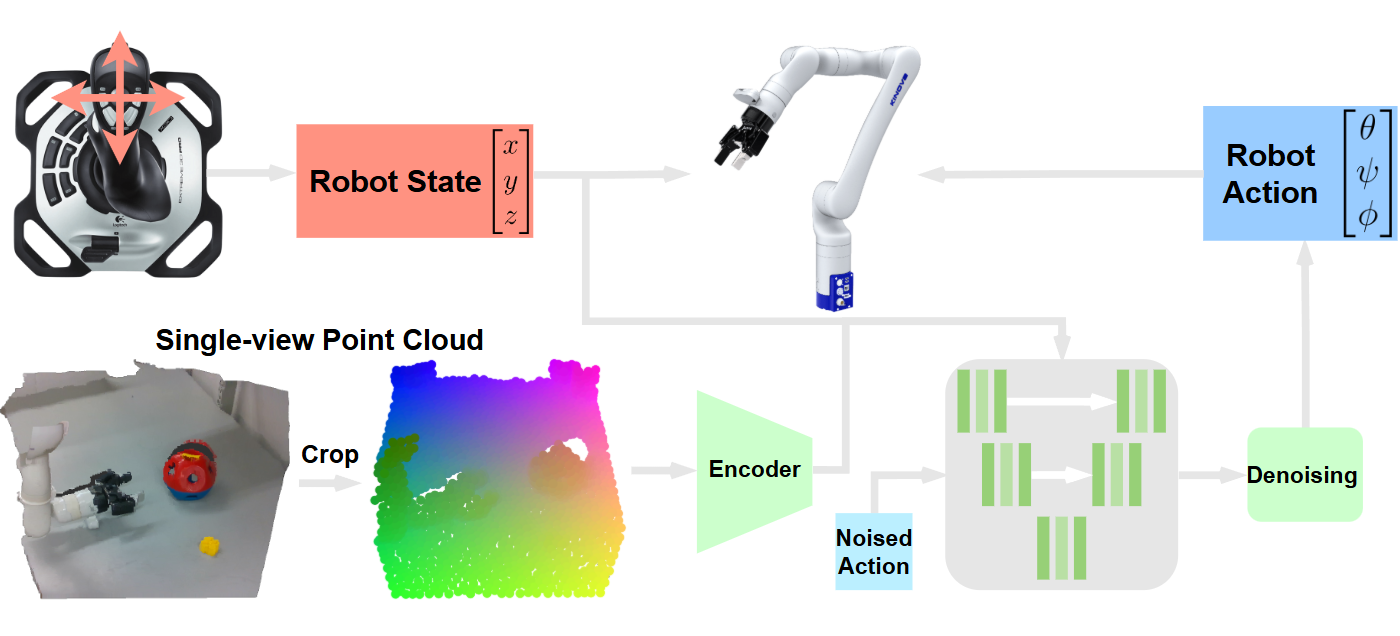}
    \caption{Overview of HITL-D methodology. During training, expert demonstrations are collected from a teleoperation interface, conditioning our robot action on a cropped and processed point cloud and robot state. During evaluation, a user will manipulate the robot by sending Cartesian position updates to the end effector. These updates run through our control loop along with point clouds to predict end effector orientations.}
    \label{methodology}
\end{figure*}

\textbf{Human In The Loop Diffusion (HITL-D)} is a purely reactive shared control approach where a diffusion policy conditions on sparsely sampled point clouds and end effector positions to predict an end effector orientation as an assisting action. This combination of human input and autonomous output allows the user to maintain control while also benefiting from non-intrusive autonomous assistance.

Orientation was chosen as the action based on two considerations: safety and workload. The user always maintains primary control of the robot as the autonomous predictions cannot change the position of the end effector. Delegating orientation to the autonomous model simplifies system use, allowing the robot to perform fine, task-specific adjustments while the human focuses on coarse positional control. This is especially beneficial for individuals with physical impairments who struggle with fine manipulation. 

In HITL-D, our diffusion policy maps observations $o \in O$ to an assisting action $a \in A$ while the user remains in primary control of the robot. Composing our system, there are 3 key parts: (a) \textbf{Autonomous Policy Training}, built upon a comparably small number of demonstrations and the perception-decision framework in \cite{ze20243d}, the action prediction is absolute end effector orientation in radians; (b) \textbf{Reactive Diffusion Policy}, adjusting certain hyper-parameters ensures the policy reacts intuitively and preserves user control in real time; (c) \textbf{Human-Diffusion Interaction}, an assistive system is developed that reacts to the users motion using techniques that enable rapid, intuitive inference, allowing the system to react predictably to user input. Figure \ref{methodology} illustrates the overall methodology and control loop.

\subsection{Autonomous Policy Training}

The autonomous component of HITL-D is a diffusion policy trained to map multi-modal observations into end effector orientation commands. 

Each training step consists of three primary input components:
\begin{itemize}
    \item \textbf{Visual}. HITL-D conditions on a colored point cloud to convey 3D spatial information, improves generalization and reduces demonstration requirements \cite{ze20243d}. The point clouds are cropped with a hand picked bounding box based on the location of the camera relative to the scene. After cropping, farthest point sampling \cite{qi2017pointnet} is performed on the point cloud reducing the number of points down to 2048. This number balances inference speed and scene representation quality. The down sampled point clouds are subsequently encoded into a 128 dimension vector for simplified and faster training using the same 3 layer MLP in \cite{ze20243d}.

    \item \textbf{Robot State}. HITL-D also conditions on a robot state vector containing the 3 dimensional x-y-z position of the end effector relative to the Robot's base frame. 

    \item \textbf{Robot Action}. HITL-D predicts a 3 dimensional end effector orientation in radians.
\end{itemize}

Our policy is designed as a conditional denoising diffusion model \cite{ho2020denoising, chi2023diffusion} following the same design choices and algorithm as \cite{ze20243d} made for their DP3 model. 

\begin{figure*}[!t]
\vspace{4pt}
    \centering
    \begin{tabular}{ccc}
        \includegraphics[width=0.31\textwidth]{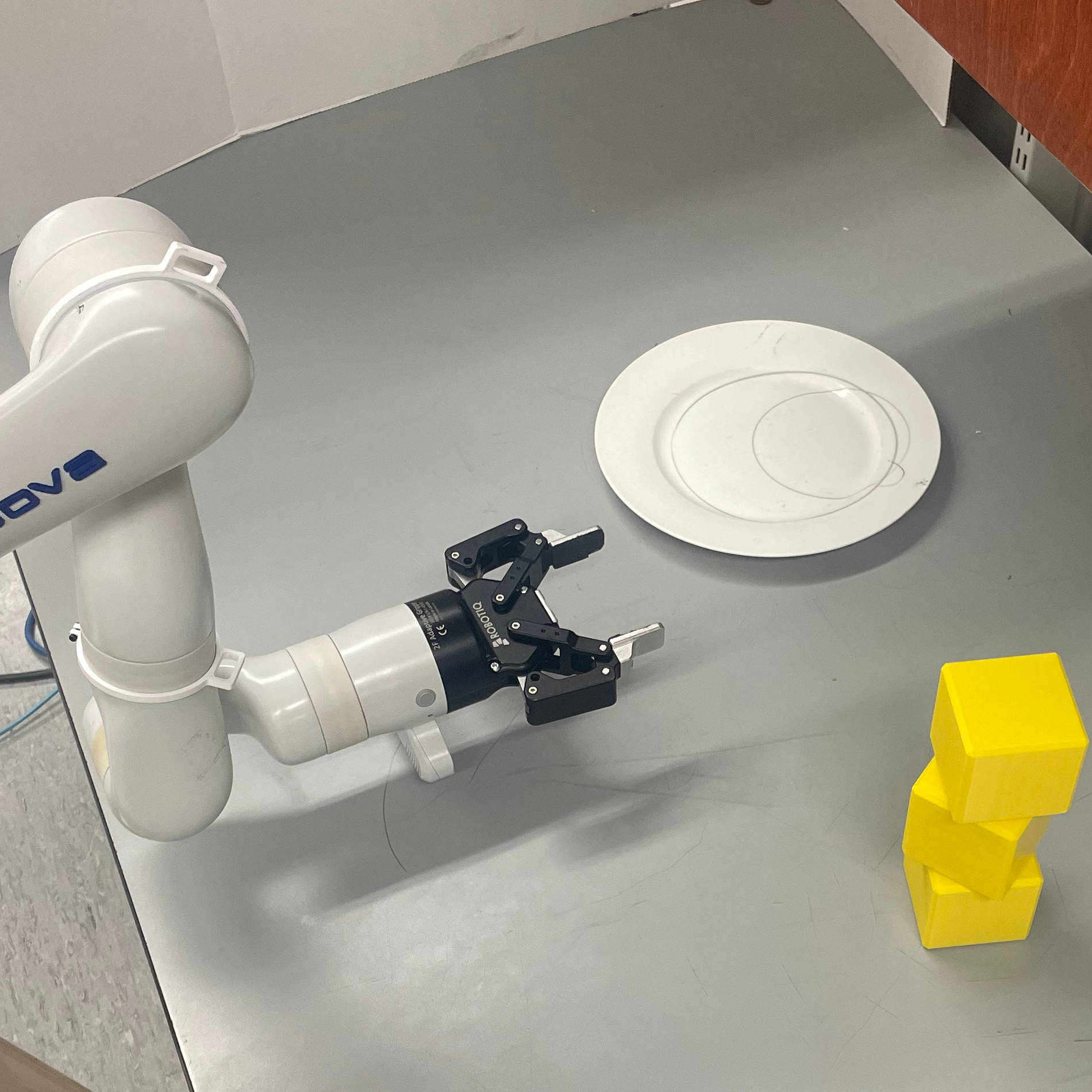} &
        \includegraphics[width=0.31\textwidth]{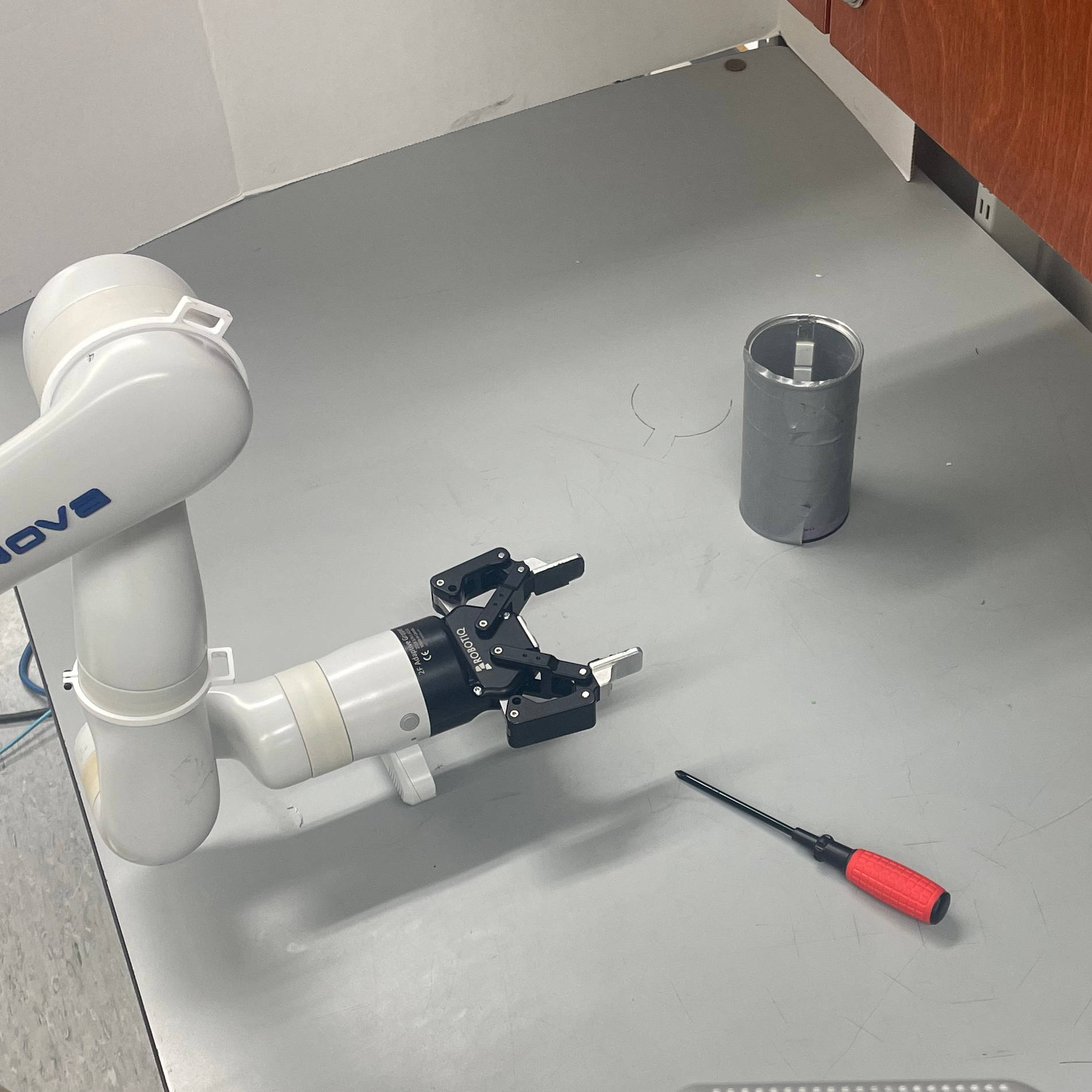} &
        \includegraphics[width=0.31\textwidth]{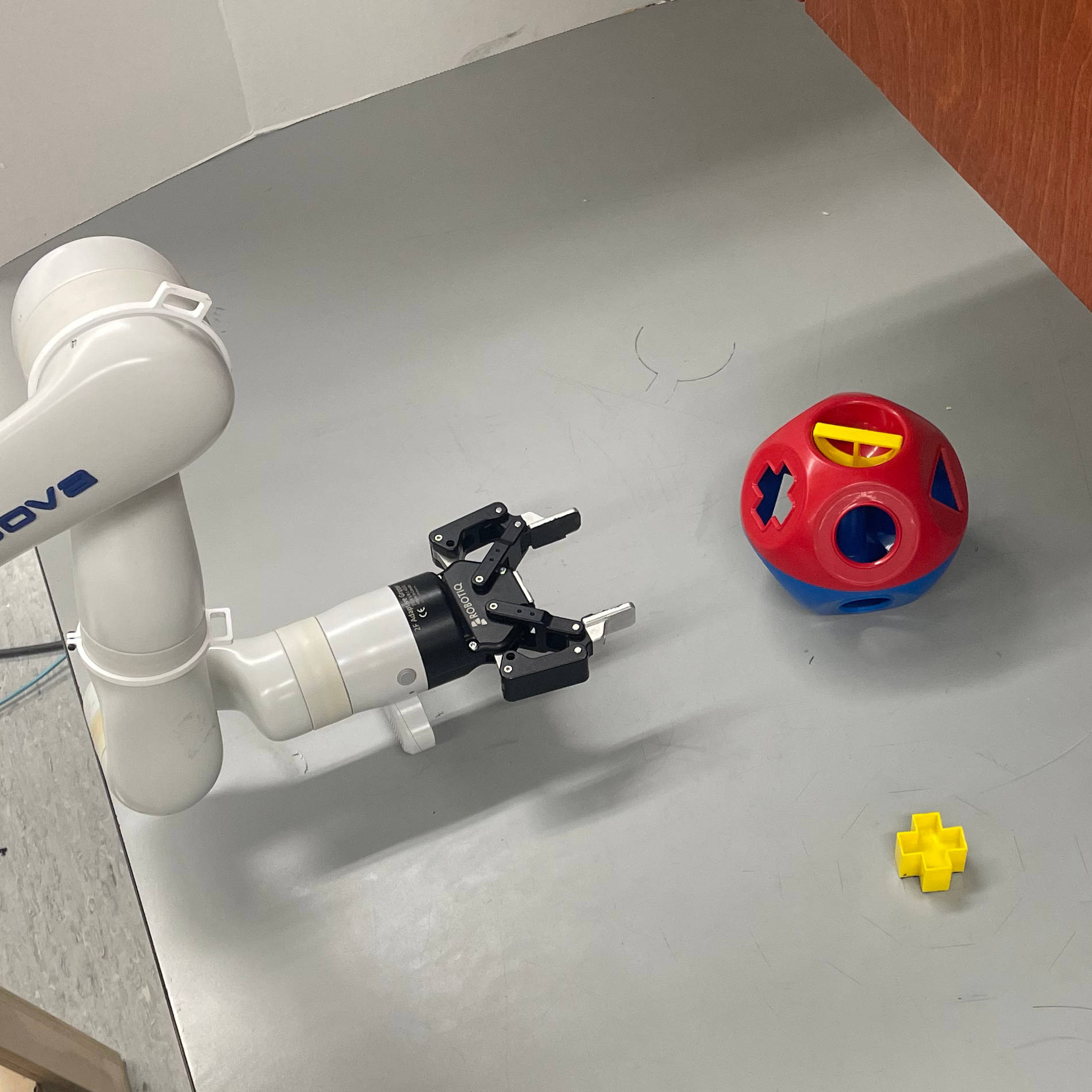} \\
        (a) Unstack Cubes Task & (b) Screwdriver Task & (c) Shape Matching Task
    \end{tabular}
    \caption{\textit{Left}: The Unstack Cubes task consists of 3 unaligned cubes and a plate, the goal is to grab each cube on their faces and place them onto the plate such that they are not stacked on the plate. \textit{Middle}: The Screwdriver Task consists of a screwdriver and a gray hollow cylinder, the goal of the task is to grab the screwdriver put it away inside the cup with the handle up. \textit{Right}: The Shape Matching task consists of a cross shape and needs to be carefully aligned with the corresponding hole in the shape matching container.}
    \label{experimentpictures}
\end{figure*}

\paragraph{Action generation}
Similar to DP3, our action is conditioned on 3D visual features $v$ and robot state vector $s$. Starting from a sample drawn from a Gaussian distribution, the model iteratively denoises this sample until convergence on a single action $a$. Let $a^K$ denote the initial noise, the denoising network $\epsilon_\theta$ performs $K$ iterations to produce a noise-free action $a^0$ according to
\begin{equation}
a^{k-1} = \alpha_k(a^k - \gamma_k \epsilon_\theta(a^k, k, v, s)) + \sigma_k \mathcal{N}(0, I),
\end{equation}
where $\mathcal{N}(0, I)$ is standard Gaussian noise, and $\alpha_k, \gamma_k, \sigma_k$ are functions of the timestep $k$ defined by the noise scheduler. This iterative procedure corresponds to the \emph{reverse diffusion process} \cite{ho2020denoising}.

\paragraph{Training objective} The network $\epsilon_\theta$ is trained to predict the noise added to a data point $a_0$ from the dataset $A$. For a given timestep $k$, the training loss is defined as
\begin{equation}
\mathcal{L} = \text{MSE}\Big(\epsilon_k, \epsilon_\theta(\bar{\alpha}_k a^0 + \bar{\beta}_k \epsilon^k, k, v, s)\Big),
\end{equation}
where $\bar{\alpha}_k$ and $\bar{\beta}_k$ are determined by the noise schedule \cite{ho2020denoising}.

\paragraph{Implementation details} HITL-D adopts a convolutional network-based diffusion policy \cite{chi2023diffusion}, using DDIM \cite{song2020denoising} as the noise scheduler. The policy is trained on 3000 epochs with a demonstration length of 256 for all of our tasks, a horizon of 1 and a previous observation size of 1. Hyper-parameter explanation is given in \ref{sec:reactive-policy}.

\subsection{Reactive Diffusion Policy} \label{sec:reactive-policy}

In pursuit of a real-time, reactive control system, we adjusted the horizon to be 1 action prediction instead of the more commonly used, 8 or 16 \cite{ze20243d, chi2023diffusion}. This prioritizes reactivity over guidance. We scrap all future predictions and just orient the end effector in response to the current Cartesian position and point cloud. Previous observation size was reduced to 1 because human motions can be unpredictable, making older observations less informative. We recorded demonstrations with sequence lengths of 256 to ensure the model observes fine-grained temporal transitions, allowing it to predict orientation corrections for a range of typical motions seen in the demonstrations.

One significant positive result of our reactive control design is that the safety is guaranteed and determined by the human's control over the robot's Cartesian position; in response, the robot may rotate its joints but it is restricted from autonomously moving to unexpected positions in the workspace, preventing risk of damage to itself or surrounding persons.

\setlength{\belowcaptionskip}{5pt}
\begin{table*}[t]
\vspace{4pt}
\renewcommand{\arraystretch}{1.2}
\caption{Quantitative results$^{1}$ from the user study comparing control systems for three tasks.}
\label{quant_results}
\centering
\setlength{\tabcolsep}{2.2pt}
\begin{footnotesize}
\resizebox{\textwidth}{!}{%
\begin{tabular}{c|ccc|ccc|ccc|ccc}
\hline
\multirow{2}{*}{\diagbox{{Experiment}}{{Metric}}} 
 & \multicolumn{3}{c|}{Completion Time (s)} 
 & \multicolumn{3}{c|}{NASA-TLX Workload (/100)}
 & \multicolumn{3}{c|}{Resets (\#/Trial)}
 & \multicolumn{3}{c}{Success Rate (\%)} \\
\cline{2-13}
& HITL-D & PnG & Cart. 
& HITL-D & PnG & Cart. 
& HITL-D & PnG & Cart.
& HITL-D & PnG & Cart. \\
\hline
Unstack     
& \bf{\underline{95.5$\pm$7.1}} & 110.4$\pm$8.0 & 124.2$\pm$9.9
& \bf{\underline{32.5$\pm$4.8}} & 42.1$\pm$5.9 & 43.8$\pm$3.8
& 0.11$\pm$0.09 & 0.06$\pm$0.04 & 0.28$\pm$0.09
& 100.0 & 91.7 & 86.1 \\

Screwdriver 
& \bf{\underline{36.5$\pm$5.8}} & 88.1$\pm$9.5 & 72.9$\pm$9.3
& \bf{\underline{25.5$\pm$3.4}} & 53.6$\pm$5.2 & 42.1$\pm$6.4
& 0.22$\pm$0.11 & 0.22$\pm$0.09 & 0.17$\pm$0.08
& 100.0 & 94.4 & 97.2 \\

Shape Match 
& \bf{\underline{56.7$\pm$6.7}} & 126.1$\pm$7.0 & 96.9$\pm$7.7
& \bf{\underline{32.0$\pm$6.0}} & 57.6$\pm$6.1 & 52.5$\pm$7.2
& 0.33$\pm$0.10 & 0.44$\pm$0.09 & 0.25$\pm$0.10
& 97.2 & 77.8 & 97.2 \\

Overall     
& \bf{\underline{62.9$\pm$19.1}} & 108.2$\pm$21.1 & 98.0$\pm$17.4
& \bf{\underline{30.0$\pm$14.5}} & 51.1$\pm$16.4 & 46.1$\pm$13.8
& 0.22$\pm$0.25 & 0.24$\pm$0.16 & 0.23$\pm$0.24
& 99.1 & 88.0 & 93.5 \\

\hline
\multicolumn{13}{l}{\scriptsize$^{1}$ Results are reported with standard error. Bold = Shared significantly better than Point and Go; Underline = Shared significantly better than Cartesian.} \\
\end{tabular}
}
\end{footnotesize}
\end{table*}

\begin{figure*}[t]
    \centering
    \setlength{\tabcolsep}{-2pt}
    \begin{tabular}{ccc}
        \includegraphics[width=0.34\textwidth]{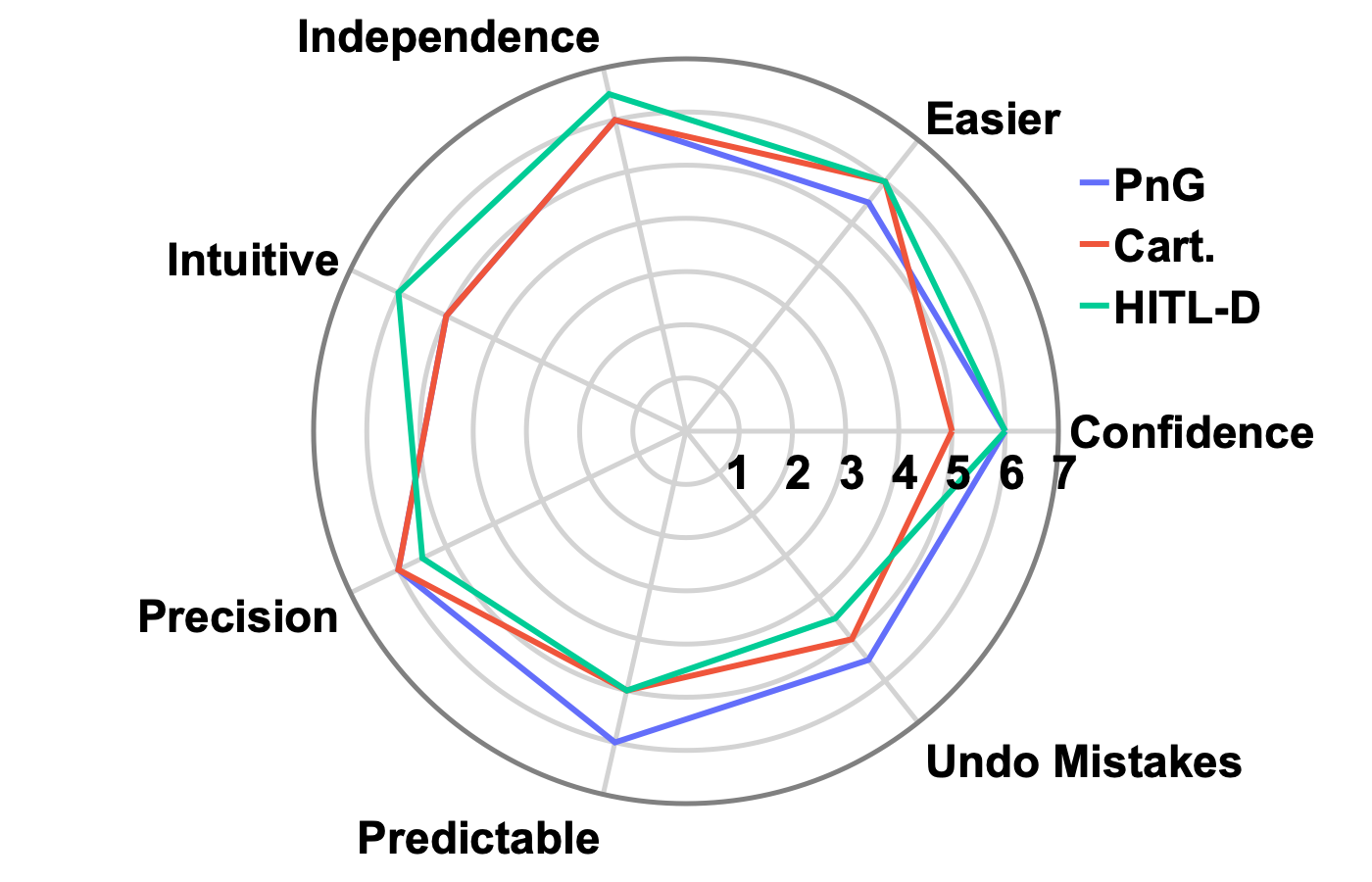} &
        \includegraphics[width=0.34\textwidth]{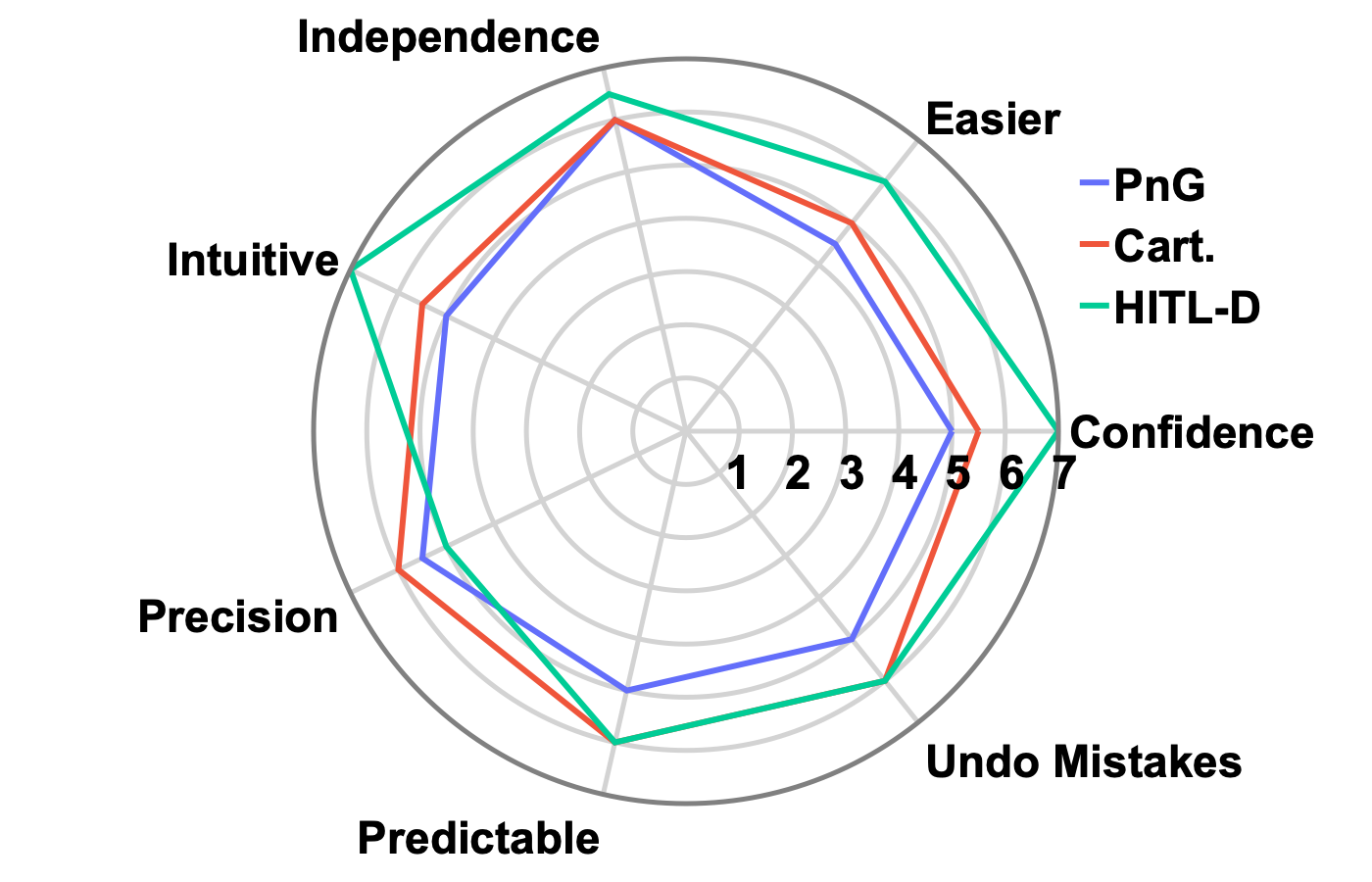} &
        \includegraphics[width=0.34\textwidth]{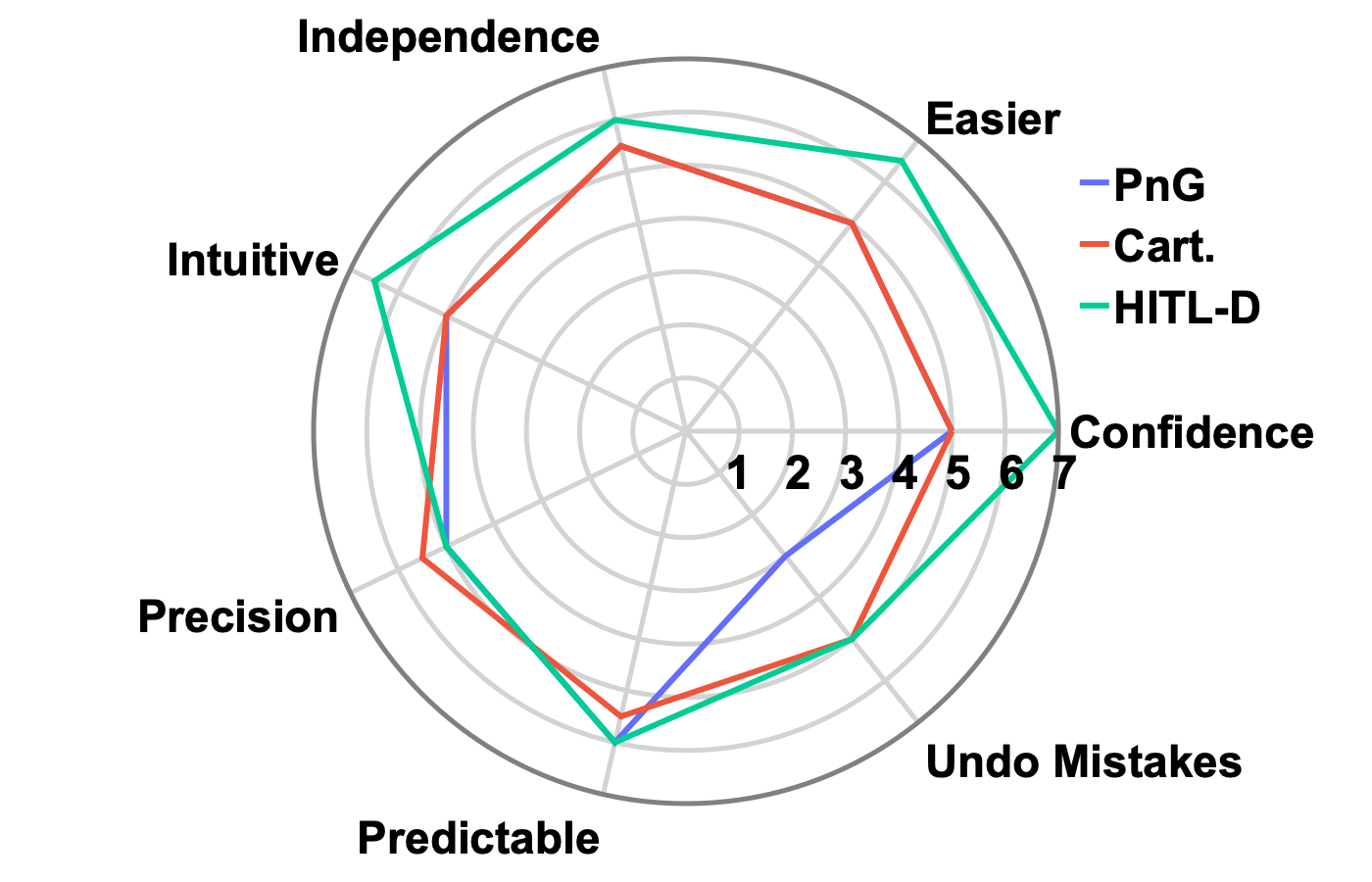} \\
        Unstack Cubes Task & Screwdriver Task & Shape Matching Task \\
        \\
    \end{tabular}
    \caption{Radar plots for Likert scale survey measuring individual ratings for different parts of the system. (1-least favorable, 7-most favorable)}
    \label{radarplots}
\end{figure*}

\subsection{Human-Autonomy Interaction}

The interaction between human control and orientation assistance via our diffusion policy is designed to be straightforward and intuitive. Since the policy predicts a desired orientation, velocity commands are generated based on the difference between the prediction and current orientations. This provides rapid corrections when large discrepancies exist, while naturally slowing movements as finer adjustments are required. The gain for this simple proportional controller is 0.05. All motion is capped by a velocity limit to maintain safety of operation. This provides a smooth and responsive change for the users while not being quick enough to feel unnatural.

\section{EXPERIMENTS}

The most important metric for evaluating the viability of our system is performance with real human users compared to other telemanipulation methods. These experiments were approved by by University of Alberta Research Ethics and Managements (Pro00054665) and signed consent was obtained from the 12 non-disabled participants (ages 19-29, 3 female). Experiments followed a single-blind format with control method randomized across trials. Each participant used each control method for each of the three tasks.
\begin{itemize}
    \item \textbf{Cartesian Control}: This method provides the user with 2 modes, one controls the 3 dimensions of Cartesian space (x, y, z) and the other mode controls end effector orientation (yaw, pitch, roll). 
    \item \textbf{Point and Go Control}: This method \cite{wang2024png} remaps the coordinate frames of the end effector creating a "point and go" behavior, it was shown that Point and Go outperformed Cartesian control is most tasks while also being better than the learned SCL method \cite{przystupa2024learning}.
    \item \textbf{HITL-D}: Our unique shared control method combining diffusion assisted orientation with human translational control.
\end{itemize}

Each participant was given a brief description of the controls and whether or not it had mode switching, but no context on what the controls did in terms of manipulating the robot. The users had 60 seconds to test and familiarize with each system before attempting to complete 3 trials for each task using each system. The three tasks used in our study (Fig. \ref{experimentpictures}) were selected to capture different types of ADLs \cite{9811960} and levels of orientation demand that average WMRA users may face during day to day activities. The unstacking task primarily tested translational control, while the screwdriver and shape matching tasks required frequent orientation adjustments. This design provided insight into whether the benefits of our shared control method extended to both orientation-intensive and orientation-light tasks.

HITL-D's diffusion policy was trained by first performing a single demonstration with an Xbox 360 controller, with Cartesian position mapped to the left stick and orientation mapped to the right stick with various functions spread across the buttons. This allowed simultaneous control over all degrees of freedom of our Kinova Gen3 Lite robot arm while doing the demonstrations. The advantage of this control setup is that our demonstrations feel smoother and more natural as there is no need to pause and switch between the translation and orientation modes. All demonstrations were performed by an expert researcher to ensure high-quality trajectories. We evaluated and trained our policy on a GeForce RTX 3090Ti.

Notably, our framework requires only one demonstration per task for action prediction, whereas autonomous baselines like DP3 \cite{ze20243d} and Diffusion Policy \cite{chi2023diffusion} utilize 40 to 284. This reduced data requirement stems from our focus on shared-control rather than fully autonomous end-to-end learning. While we do not claim broad generalization or robustness from a single sample, consistent with broader research in robotic diffusion, we expect that increasing the demonstration count would further enhance policy capability and adaptability.

\subsection{Study Protocol and Metrics}

Participants were asked to perform three tasks. For each task, they used three separate control methods across 3 trials. The robot was always in the same starting position and each scene setup was replicated for each user in the study. At the end of each task-method pair, the users were asked to fill out 2 surveys.

\begin{enumerate}
    \item \textbf{Likert Survey:} A 7 point survey measuring the following metrics;
    \begin{itemize}
        \item \textit{Confidence}: how confident are you in using the teleoperation mapping?
         \item \textit{Easier}: how easy was completing the task with this method?
        \item \textit{Independence}: how independently do you feel you can control the robot?  
        \item \textit{Intuitive}: how intuitive was the system?
        \item \textit{Precision}: how precise were the commands in the system?
        \item \textit{Predictable}: how predictable was the robots behavior?
        \item \textit{Undo}: how easy was it to undo mistakes with this system?
    \end{itemize}
    \item \textbf{NASA Task Load Index (TLX):} \cite{hart1988development} developed a system to measure subjective workload across different tasks. In this study, it is used to enable meaningful comparisons between methods.
\end{enumerate}

Trial videos, task completion time, successes and failures, mode switches, completion times and the number of resets were collected for this study. However, mode switches were not collected for our HITL-D method as there is only 1 mode. A reset for a task was required when either the cubes were knocked over or the screwdriver was not dropped in the cylinder and became unreachable. A user could also request a reset if the their control resulted in a robot configuration they could not understand or recover from.

Comparisons of quantitative metrics were achieved with two-way repeated-measures ANOVA. Binary success rates were evaluated with Cochran's Q test. For discrete data from the Likert assessments, Wilcoxon signed-rank tests were conducted. The threshold for significance was set at 0.05.

\begin{figure}[t]
    \centering
    \begin{tabular}{ccc}
        \includegraphics[width=0.9\columnwidth]{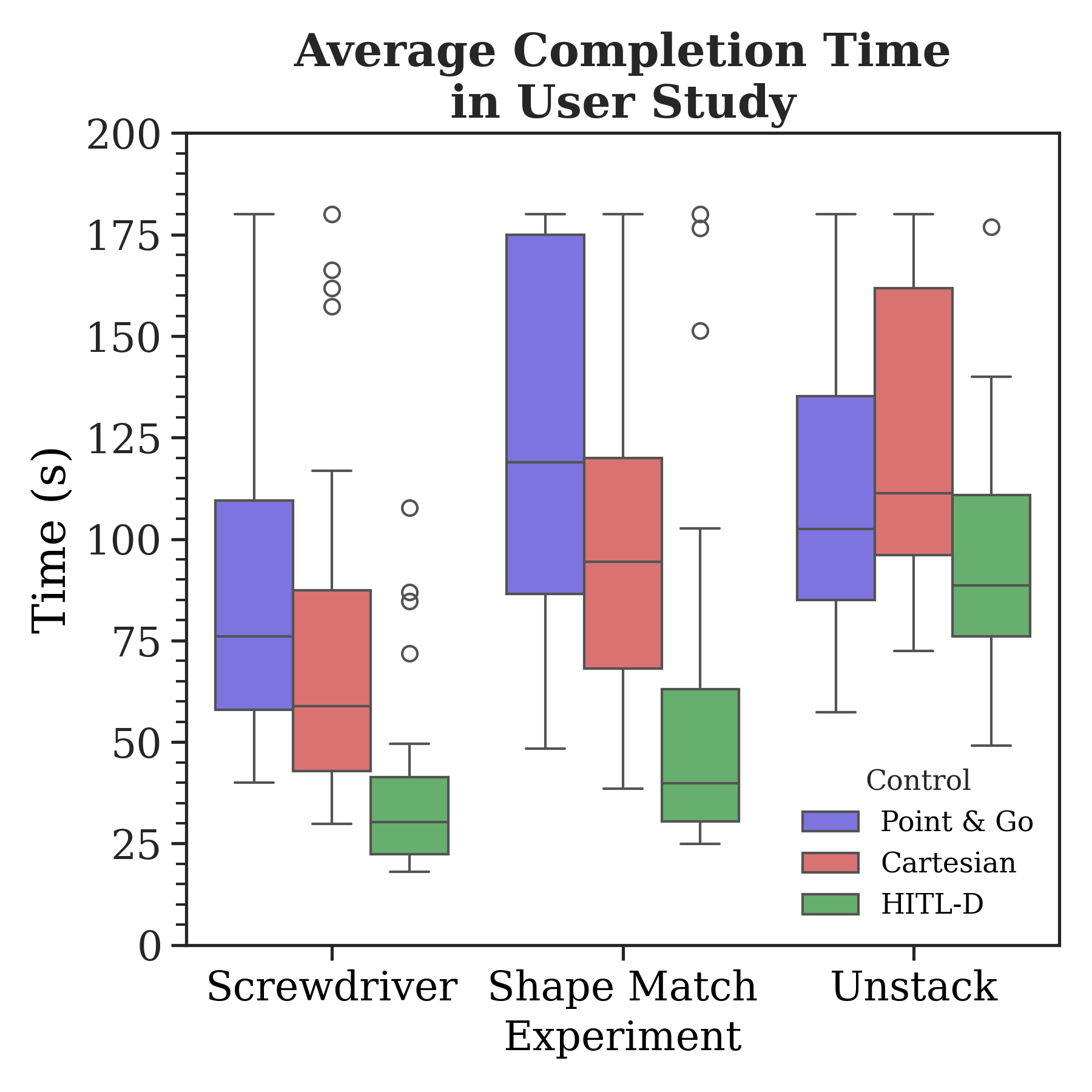} \\
    \end{tabular}
    \caption{Box plot for completion times showcasing clearer differences in means, deviations and outliers}
    \label{boxtimes}
\end{figure}

\section{RESULTS}

Table \ref{quant_results} shows completion times, NASA-TLX workload scores, number of resets and success rates for each task-method pairing. Figure \ref{radarplots} displays the Likert scale metrics for both robot behavior and perceived competence. 

\begin{figure}[t]
    \centering
    \begin{tabular}{ccc}
        \includegraphics[width=0.9\columnwidth]{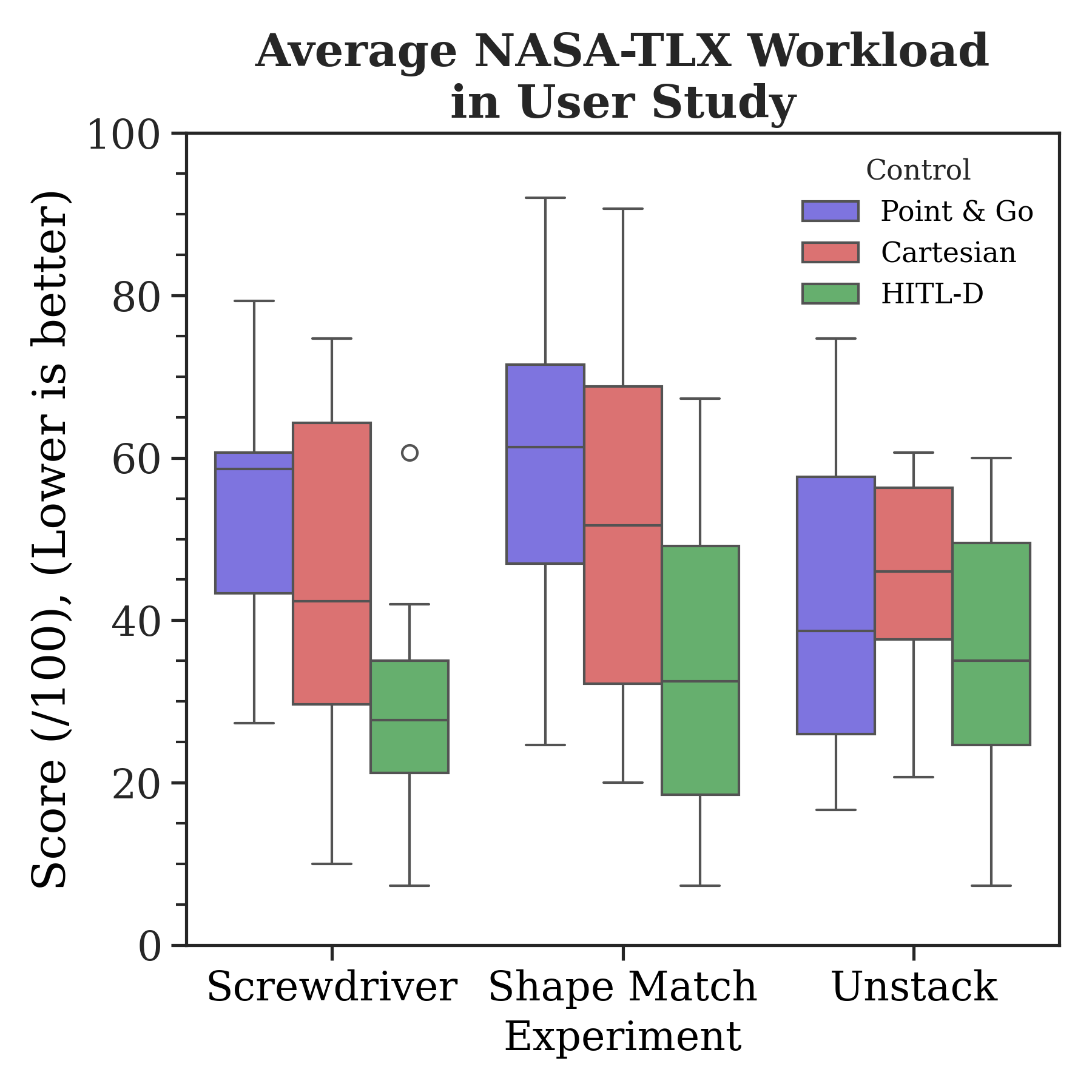} \\
    \end{tabular}
    \caption{Box plot for NASA-TLX Workload ratings showcasing clearer differences in means, deviations and outliers}
    \label{boxworkloads}
\end{figure}

Overall we observed a significant decrease in completion times and perceived workloads across all tasks with HITL-D. Notably, standard deviations were lower for each metric with the exception of average resets per trial, indicating that performance improvements were consistent across participants. Figures \ref{boxtimes} and \ref{boxworkloads} demonstrate the visual significance of HITL-D's improvements over Cartesian and Point and Go.

\textbf{Unstack Task:} HITL-D showed significant improvements with a time reduction of 13.5\% and a NASA-TLX workload improvement of 22.8\% in comparison to Point and Go. HITL-D was 23.1\% faster and had a 25.8\% lower workload than Cartesian. HITL-D maintained competitive in the Likert scores of independence and predictability and while not statistically significant, HITL-D showed trends of being more intuitive to the users. We also observed that HITL-D had a 100\% success rate while Point and Go and Cartesian methods did not.

\textbf{Screwdriver Task:} We observe significant improvements of HITL-D over both PnG and Cartesian methods in completion times and NASA-TLX workload scores. HITL-D was nearly twice as fast as Cartesian, reducing average completion time by 49.9\%, the workload score was 39.4\% lower. HITL-D was more than twice as fast as PnG improving the average time by 58.6\% and more than halving the workload score by with an improvement of 52.4\%. HITL-D again maintained a success rate of 100\% while Cartesian and PnG were close by at above 94\% success rates in all trials. HITL-D also showed higher scores, and while the differences were not statistically significant, nearly all Likert survey metrics, except precision, indicate that users preferred HITL-D for this task.

\textbf{Shape Match Task:} Again, we observe significant improvements of HITL-D over both PnG and Cartesian methods in completion times and workload scores. In comparison with Cartesian, HITL-D reduced average completion time by 41.2\%, the workload score was 39.0\% lower. HITL-D was more than twice as fast as PnG improving the average time by 55.0\% and reducing the workload score by 44.4\%. HITL-D had a very high success rate of 97.2\% and while the PnG struggled to keep up at 77.8\% success rate, Cartesian tied HITL-D at 97.2\%. HITL-D improved in most metrics in the Likert scale survey except precision which still matched the other two methods. 

\section{DISCUSSION}

\textbf{Unstack Task:} The comparatively smaller performance gain of HITL-D over Cartesian and PnG methods in the unstacking task can be explained by the fact that this task does not require large or frequent orientation changes, depending on the user’s approach. We observed that some users bumped or pushed the middle cube to rotate it into a position they could grasp while others spent the time switching modes to get the correct orientation. The statistically significant improvement provided by HITL-D arises from its ability to automatically orient with respect to the cubes correctly, skipping the bumping step and reorienting toward the plate. We make the claim that our method is at least as effective as traditional teleoperation methods. Supporting this claim, the results from the unstack cubes task demonstrate that in situations where orientation adjustments aren't necessary, our method maintains or improves almost all metrics measured in the Likert survey of our method in comparison to the other two. 

\textbf{Screwdriver Task:} Our evaluation indicates that users significantly benefited from the shared control framework in this task. Some users picked up the screwdriver completely perpendicular to the table and others picked it up at quite a shallow orientation before the autonomous component could reach its prediction. Despite having only one demonstration, the model was still capable of assisting the user effectively with variations in user strategy leading to a 100\% success rate. HITL-D's autonomous component for orientation proved significantly useful in user workload ratings as well as the empirical evidence in completion times.
We noticed that the areas users spent the most time in while using Point and Go was when the forearm joints needed to flip due to the change of control frames, which lead to visible frustration. Users who did not hit this case of joint flipping were not as frustrated and were able to complete the task with competitive times.
The Cartesian control method also proved to be complex for users who did not use the mode switching for the Unstack task before this one. They were not as familiar with the orientation mode, causing them to perform worse. 
Users typically failed at this task or needed resets due to dropping the screwdriver outside of the tube. One interesting set of trials during the study led to a false negative for the HITL-D method, during the Cartesian experiment the user would lightly pick up the screwdriver so that it had enough friction to stay in the grippers but just low enough so that it would rotate within them to be completely vertical. When they were unable to get this technique working with HITL-D they assumed that it was the control method that wasn't doing what they desired and kept attempting this behavior and needing to reset. Figure \ref{boxtimes} demonstrates these times as outliers as they were the only inconsistent trials during the experiment. Subsequently this false negative resulted in the only outlier for the workload rating seen in Figure \ref{boxworkloads}. Even with this outlier HITL-D remains significantly well suited for a task like this that necessitates large orientation changes.

\textbf{Shape Match Task:} This task posed the greatest difficulty for users, as achieving a successful insertion required the end effector orientation to be nearly exact in order to align with the shape-matching container. Users commonly struggled with the roll of the end effector for the Cartesian method which caused the cross to be unaligned. Similar to the screwdriver task, users would pick up the cross at a very shallow angle, resulting in them not having enough pitch available in the workspace to orient the end effector so the cross had the right orientation. We also noticed a particular behavior only for the HITL-D method due to its shared control nature. Users inexperienced with the method and task often attempted to pick up the plus shape at a shallow angle. Upon trying to insert it they found that they were constrained by the end effectors ability to orient correctly with various pick up orientations and would have to reset. Evidence for this behavior is demonstrated in Figure \ref{boxtimes} which visualizes the outliers in trial times. Once users recognized that the autonomous model required an additional period to achieve an optimal orientation when picking up the cross, they were able to complete the task more quickly, and no resets were needed after the first trial.

\textbf{Across All Tasks:} Overall, the observations demonstrate the efficacy of our method in time to completion, perceived workload, and on average higher user preferences for our method in the Likert metrics such as intuitiveness, ease of control and independent ability to move the robot. An interesting result we observed, although not statistically significant, was that Point and Go showed trends in higher task completion times and workload ratings than Cartesian in the Screwdriver and Shape Match tasks. We attribute this to the fact that Point and Go is a recent novel control method \cite{wang2024png} and that while it may require more time to learn, it shows promising results for experienced users. Most importantly, we proved our claim that our method does not reduce human autonomy: users experienced increased autonomy and independence compared to that of Point and Go and Cartesian methods.

\section{CONCLUSION}

In this work, a novel combination of diffusion and human input for robot teleoperation is proposed, where a user controls a robot via a joystick and sends Cartesian velocity commands and a diffusion policy reacts to the Cartesian pose to provide an end effector orientation to assist the user in a reactionary setting. Through a user study, users maintained a sense of independent control while experiencing a significant reduction in mental workload using our framework. We observed statistically significant improvements in task completion, highlighting the efficiency and usability of our approach, especially with reduced-DoF controllers. These findings emphasize the benefits of shared control over purely human or autonomous control, demonstrating the unique advantages of combining diffusion-based fine manipulation capabilities with human task expertise. Future work includes conducting a user study on impaired individuals to validate the usability and workload reduction for our intended demographic. Additionally, evaluating a wider variety of ADLs would give better insight into the applicability of this method. Further, we acknowledge the limitation of a hand cropped point cloud and want to explore additional ways to improve invariance to object, camera and robot arm variations. Another direction is incorporating visual recognition models to better localize objects in the scene, which could enable benefits like dynamic cropping and new forms of conditional data. 


\bibliographystyle{IEEEtran}
\bibliography{references}


\end{document}